\documentclass[sn-mathphys-num]{sn-jnl}

\usepackage{graphicx}%
\usepackage{subfig}
\usepackage{multirow}%
\usepackage{amsmath,amssymb,amsfonts}%
\usepackage{amsthm}%
\usepackage{mathrsfs}%
\usepackage[title]{appendix}%
\usepackage{xcolor}%
\usepackage{textcomp}%
\usepackage{manyfoot}%
\usepackage{booktabs}%
\usepackage{algorithm}%
\usepackage{algorithmicx}%
\usepackage{algpseudocode}%
\usepackage{listings}%

\theoremstyle{thmstyleone}%
\theoremstyle{thmstyletwo}%

\theoremstyle{thmstylethree}%

\begin{document}

\title[Article Title]{Semantically Robust Unsupervised Image Translation for Paired Remote Sensing Images}


\author[1]{\fnm{Sheng} \sur{Fang}}\email{fangs99@126.com}

\author[2]{\fnm{Kaiyu} \sur{Li}}\email{likyoo.ai@gmail.com}

\author*[1]{\fnm{Zhe} \sur{Li}}\email{lizhe@sdust.edu.cn}

\author[1]{\fnm{Jianli} \sur{Zhao}}\email{jlzhao@sdust.edu.cn}

\author[1]{\fnm{Xingli} \sur{Zhang}}\email{zhangxl@sdust.edu.cn}

\affil*[1]{\orgdiv{College of Computer Science and Engineering}, \orgname{Shandong University of Science and Technology}, \orgaddress{\city{Qingdao}, \postcode{266590}, \country{China}}}

\affil[2]{\orgdiv{School of Software Engineering}, \orgname{Xi'an Jiaotong University}, \orgaddress{\city{Xi'an}, \postcode{710049}, \country{China}}}


\abstract{Image translation for change detection or classification in bi-temporal remote sensing images is unique. Although it can acquire paired images, it is still unsupervised. Moreover, strict semantic preservation in translation is always needed instead of multimodal outputs. In response to these problems, this paper proposes a new method, SRUIT (Semantically Robust Unsupervised Image-to-image Translation), which ensures semantically robust translation and produces deterministic output. Inspired by previous works, the method explores the underlying characteristics of bi-temporal Remote Sensing images and designs the corresponding networks. Firstly, we assume that bi-temporal Remote Sensing images share the same latent space, for they are always acquired from the same land location. So SRUIT makes the generators share their high-level layers, and this constraint will compel two domain mapping to fall into the same latent space. Secondly, considering land covers of bi-temporal images could evolve into each other, SRUIT exploits the cross-cycle-consistent adversarial networks to translate from one to the other and recover them. Experimental results show that constraints of sharing weights and cross-cycle consistency enable translated images with both good perceptual image quality and semantic preservation for significant differences.}




\maketitle

\section{Introduction}\label{sec1}

Over the past decade, image translation methods have emerged as a significant area of research, garnering increasing attention from diverse communities, including medicine and remote sensing (RS). By leveraging the multiple levels of representation learned by deep neural networks\cite{Lecun:dl}, these methods enable the generation of realistic images in different domains from a given set of images. The success of generative models, such as GAN\cite{Goodfellow:NIPS2014}, VAE\cite{Kingma:ICLR2014}, or VAE-GANs\cite{Rezende:ICML2015}, has further fueled the development of various image translation methods for different applications.

According to supervision, image-to-image translation methods could be coarsely divided into two types. The first is paired image translation\cite{Isola:CVPR2017,Hoffman:ICML2018,Zhu:NIPS2017}, which uses the paired images from the source and target domain to supervise each other to train the translation model. The need for strict one-to-one input data heavily limits its application. In contrast, the second type of unpaired image translation\cite{Zhu:ICCV2017,Liu:NIPS2017,Huang:ECCV2018} uses domain-set image statistics as supervision. It means that unpaired image translation is unsupervised at the image level. This relaxes paired images' strict demand and leads to more convenient applications. In general, using supervision flexibly is the critical key to image translation.

\begin{figure}[h]
  \vspace{-0.1cm}
  \centering
  \subfloat[]{\includegraphics[width=0.25\linewidth]{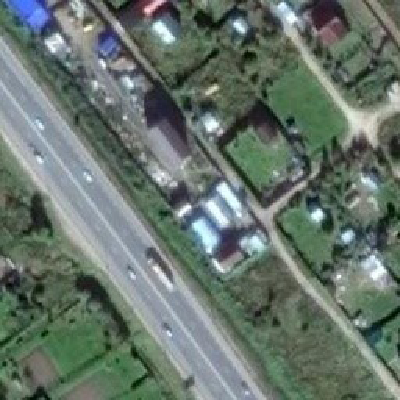}
  \label{subfig_f1a}}
  \subfloat[]{\includegraphics[width=0.25\linewidth]{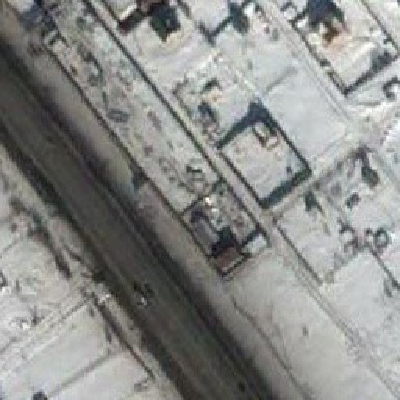}
  \label{subfig_f1b}}
  \subfloat[]{\includegraphics[width=0.25\linewidth]{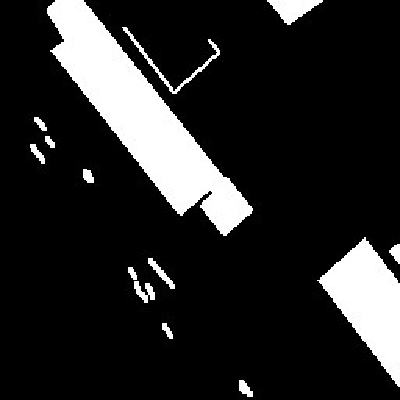}
  \label{subfig_f1c}}
  \caption{Bi-temporal RS images with season-varying in CDD dataset\cite{Lebedev:ISPRS2018}, (a) and (b) are the original bi-temporal images taken in summer and winter, and (c) is the ground truth map of changed areas.}
  \label{fig1}
\end{figure}

In recent years, most works have focused on unpaired image-to-image translation. Nevertheless, in some applications, the question is different. For example, in the change detection of RS images, paired images are always at hand but have different translation demands. The bi-temporal RS images, acquired at the same land locations at different times, are compared to find changed areas, as Fig. \ref{fig1} shows. It is difficult because there are significant appearance differences between bi-temporal RS images, which will heavily hinder the task. Although recent developments in change detection methods based on deep learning have improved detection accuracy\cite{Fang:SNUNET, Fang2023Changer, Zhang:DSIFNET}, they are still haunted by this question.

One reasonable way to alleviate the problem is to translate the summer image as Fig. \ref{subfig_f1a} to the winter image as Fig. \ref{subfig_f1b} while keeping its semantics. So, it brings out a new question. Because some land cover areas have changed and we have no access to the appearance of changed areas in the target domain, translation between bi-temporal RS images with remarkably different appearances is unsupervised. However, the paired images are at hand. So, how to fully use the paired images to implement unsupervised translation should be dealt with. The core of the question is how to preserve the semantics of changed areas. For a typical paired image-to-image method, e.g., Pix2Pix\cite{Isola:CVPR2017}, the $L1$ loss function between the source and translated images will compel the image to approximate the source. This is not suitable for the above question. The cycle-consistent method, e.g.\cite{Zhu:ICCV2017}, will be competent for this task. However, the real-like translated images refer to a similar distribution instead of semantics\cite{Fu:CVPR2019,Jia:ICCV2021,Yang:CVPR2020}, and parts of its content have different semantics from the source. It is more difficult to generate high-quality images. Although many works\cite{Gong:GAN,Gong:GDCNET,Jose:GARSS2020,Luppino:TNNLS2024} utilize generative models to implement the translation between bi-temporal images, they often train the model in the unchanged areas\cite{Gong:GAN,Gong:GDCNET,Jose:GARSS2020}, which are selected through other methods or align the latent space with extra information. As we know, seldom works of image-to-image translation study this subject.

This paper proposes a new method, SRUIT (Semantically Robust Unsupervised Image-to-image Translation), for RS images in response to these problems. It is designed based on two observations and corresponding assumptions. First, bi-temporal RS images are acquired from the same land location, so we assume the paired images share the same latent space. The second is that bi-temporal images represent the same land location at different times so that they could be translated into each other in bi-direction. It consists of two generators; one is responsible for translating samples from domain $\mathscr{A}$ to domain $\mathscr{B}$, and the other vice versa. The weights of high-level layers of two generators are shared to enforce the projection of bi-temporal images into the same latent space. Further, the cross-cycle consistency constraint ensures semantic preservation in translation instead of cycle consistency.

We conduct a comprehensive quantitative and qualitative comparison of SRUIT with several typical image translation methods. The experiments on diverse datasets demonstrate that SRUIT can implement semantically robust unsupervised translation and is helpful for the downstream task.
Our contributions are summarized as follows:
\begin{itemize}
\item We clarify that some image translations, typical as bi-temporal RS images used in change detections or classifications, differ from previous studies\cite{Isola:CVPR2017,Hoffman:ICML2018,Zhu:NIPS2017,Zhu:ICCV2017,Liu:NIPS2017,Huang:ECCV2018}. Although we have access to the paired images, they are still unsupervised. More important, the changed areas must preserve their semantics during translation.
\item Based on the characteristics of bi-temporal RS images, SRUIT exploits constraints of sharing weights in high-level layers and cross-cycle consistency to implement the unsupervised image translation with semantics preservation.
\item Experimental comparisons prove our method can produce good perceptual quality images with semantic preservation. In addition, no metrics are suitable for measuring translated images both in semantic preservation and image perceptual quality until now, and further research is needed. Code and pre-trained models are available at https://github.com/xxxxxx.
\end{itemize}

\section{Related Works}\label{sec2}
\subsection{Generative Model}\label{sec2.1}
Generating real-like images has been one of the essential topics in computer vision. In the last decade, some important works have been proposed according to various philosophical ideas, typical as using adversarial networks \cite{Goodfellow:NIPS2014}, approximating the actual distribution by variational auto-encoder\cite{Kingma:ICLR2014}, and applying direct mapping between source and target distributions\cite{Rezende:ICML2015}. More details about these generative models can be referred to \cite{Xia:TPAMI2023,DBLP:Tschannen2018,Ivan:NF}. These outstanding works and their varieties significantly promote vital studies of image translation. Primarily, we are inspired by the ideas of \cite{Liu:NIPS2017,Liu:NIPS2016,Yusuf:CMSNET}, which generate the translated images from a shared latent space using a weight-sharing constraint. Unlike them, which try to generate multimodal outputs, our method aims to produce deterministic output depending on the paired images.

\subsection{Image-to-Image Translation}\label{sec2.2}
Pix2Pix\cite{Isola:CVPR2017} and Cycle-GAN\cite{Zhu:ICCV2017} will be briefly introduced and taken as the basis of supervision and semantic preservation. Because almost all works use GAN loss as a part of their objective, we will focus on the extra loss and constraints.

Pix2Pix uses the paired images to train the model, and the $L1$ loss between translated and actual images is taken as an extra constraint. This compels translated images in the same way as the target, and skip connections of the generator enhance this effect at the pixel level. In addition, the noise vector is a dropout, so its capacity to produce diverse outputs is limited. Aiming at the above questions, Cycle-GAN uses domain-set images instead of paired images to train two translation models jointly. The cycle loss is exploited to incentivize mapping an individual image to the desired output.

Considering combining both advantages of Pix2Pix and Cycle-GAN, Cycada\cite{Hoffman:ICML2018} uses paired images as input and explores cycle-consistent adversarial networks. Moreover, it uses an extra classification task to aid semantic preservation. Unlike Cycle-GAN, UNIT\cite{Liu:NIPS2017} also uses the cycle translation but divides its networks into the parts of VAE and GAN and uses a weight-sharing constraint between two VAE models in their high representation layers. In addition to the loss of VAE and GAN, an extra cycle loss based on KL divergence is computed to ensure the latent space alignment. However, according to the reports of UNIT, the sharing weights contribute little to the performance compared to using cycle-consistence alone. According to our experiments, it is mainly because the images from two domains may deviate too much in latent space. Moreover, UNIT appends one noise vector to a feature vector extracted by the encoder to produce diverse outputs. Discarding the sharing weights constraint of UNIT, MUNIT\cite{Huang:ECCV2018} uses two networks of content and style encoder to produce multimodal outputs, and the corresponding loss items of content and style are used as extra constraints. In addition, more techniques, such as attention\cite{Mejjati:NIPS2018}, adaptive normalization\cite{DBLP:Kim2019}, disentangled representations\cite{Lee:DRITplusplus}, and marginal prior knowledge\cite{You:BCCGAN}, are used to improve translation performance.

In general, the above works show that cycle consistency is helpful for individual images to maintain their semantics. However, it often becomes invalid, as \cite{Fu:CVPR2019,Jia:ICCV2021,Yang:CVPR2020,Gong:GAN} reported. The reason may be that cycle consistency constraints cannot fully alleviate the difference between two distributions of images or domains. Moreover, diversity and semantic preservation are contrary goals for unsupervised image translation. Many works\cite{Zhu:ICCV2017,Liu:NIPS2017,Huang:ECCV2018,Mejjati:NIPS2018,DBLP:Kim2019,Lee:DRITplusplus,You:BCCGAN} try to achieve both goals concurrently. Semantic preservation is essential for applications such as remote sensing and needs further research. Unlike the above methods, our method tries to generate the deterministic output and let its semantics be the same as the source.

\subsection{Style Transfer}\label{sec2.3}
Image-to-image translation has strong relations with style transfer. The work of \cite{Gatys:CVPR2016} firstly applied CNN to style transfer, and then many works focused on normalization layers to carry out style transfer. AdaIN\cite{Huang:ICCV2017} and its varieties are often integrated into image translation\cite{Huang:ECCV2018,DBLP:Kim2019,Park:CVPR2019,Jiang:ECCV2020}. In MUNIT\cite{Huang:ECCV2018}, style encoders extract style features of a specific style image and apply them to the content image in AdaIN. Aiming at the question of the normalization layers of 'wash away' semantic information, SPADE\cite{Park:CVPR2019} integrates AdaIN into the Pix2Pix model. It uses a CNN network to extract style information from a pre-defined mask. In \cite{Jiang:ECCV2020}, the style and content features from the style and images are applied in each layer to implement the feature adaptive normalization and denormalization, respectively. These methods fit for the translation of RS images. However, balancing the style and content features in translated images is difficult. Experiments show that strange textures often haunt the transferred images, and some areas are changed.

\subsection{Semantic Robust Image Translation}\label{sec2.4}
Although more attention is paid to producing real-like multimodal outputs in image translation fields, preserving semantics still attracts lots of attention. Unlike Cycle-GAN, which uses cycle consistency, Distance-GAN\cite{Benaim:NIPS2017} uses an extra loss of absolute difference between the distances of paired images in each domain. If there is no access to paired images, an alternative of self-distance can be employed\cite{Benaim:NIPS2017}. Distance-GAN is especially fit for one-sided domain mapping. In \cite{Fu:CVPR2019}, some geometry transformations are applied to the images. The original image and its transformations are all translated into the target domain. Extra loss named geometry consistency is computed between the translated image and its inverse transformations. Like \cite{Benaim:NIPS2017}, it is also used for one-sided domain mapping and performs well in preserving semantics. However, good performance is at the cost of extra models. Each transformation needs one discriminator. Inspired by visual psychophysics, \cite{Yang:CVPR2020} integrates the phase consistency constraints into the domain mapping for semantic preservation. However, our experiments show that only using the phase-consistency constraint in RS images may lead to artifacts in the faked image. In addition, \cite{Jia:ICCV2021} studies the question of semantic robustness under the frame of contrastive learning\cite{Park:ECCV2020}, and semantic robustness is acquired by making the semantics of translated output invariant to slight feature space variations of the inputs.

Our method is different from the above works in three aspects:
\begin{itemize}
\item We aim at unsupervised domain mapping with access to paired images. In order to fully utilize this advantage of paired images, SRUIT exploits the shared-weights constraint. 
\item A cross-cycle consistency constraint ensures semantic preservation for the paired images from the same land location and shares latent space. 
\item Our method jointly trains target and source domain generators and demands no other networks and tasks as help.
\end{itemize}

\section{Method}\label{sec3}
\subsection{Assumptions}\label{sec3.1}
\begin{figure}[h]
  \centering
  \includegraphics[width=0.45\linewidth]{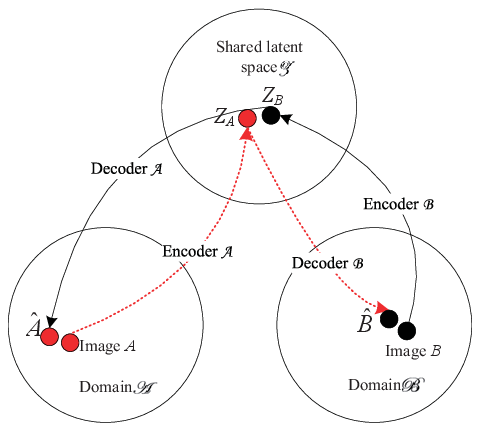}
  \caption{Illustration of bi-temporal image translation from the source domain to latent space and then to the target domain when some land cover areas are changed between images $A$ and $B$. If they have the same land covers, the translated sample should fall into the same point with the target in both shared latent space and target domain.}
  \label{fig2}
\end{figure}

Let $A\in\mathscr{A}$ and $B\in\mathscr{B}$ be image patches from two domains, e.g., bi-temporal RS images. In our unsupervised image translation setting, we have no access to the joint distribution $p(A, B)$ while two marginal distributions $p(A)$ and $p(B)$ from which the samples were drawn are given. For change detection, its final goal is to estimate the conditional distribution $p(A|B)$ or $p(B|A)$. It is a challenging task to directly estimate if bit-temporal images have significant differences in appearance, as Fig. \ref{fig1} shows. Fortunately, we can estimate the two distributions through the translation models $p(\hat{B}|B)$ and $p(\hat{A}|A)$, where $\hat{A}$ and $\hat{B}$ are samples produced, as shown in Fig. \ref{fig2}. Unlike the data used in previous works\cite{Isola:CVPR2017,Hoffman:ICML2018,Zhu:NIPS2017,Zhu:ICCV2017,Liu:NIPS2017,Huang:ECCV2018,Baek:ICCV2021,Lebedev:ISPRS2018,Fu:CVPR2019,Jia:ICCV2021,Yang:CVPR2020,Liu:NIPS2016,Yusuf:CMSNET,Mejjati:NIPS2018,DBLP:Kim2019,Lee:DRITplusplus,You:BCCGAN}, in the cases of change detection, paired images are always at hand, but between them, changes in land covers may happen, as Fig. \ref{fig1} shows. This observation means that although paired images have different appearances, the contents represented by paired images are naturally the same in many areas. Based on this observation, we assume two domains, $\mathscr{A}$ and $\mathscr{B}$, share the same latent space $\mathscr{Z}$. Even if some changes happened, they still fall into the neighborhood of each other in the shared latent space, as Fig. \ref{fig2} shows.

\begin{figure}[h]
  \centering
  \includegraphics[width=0.45\linewidth]{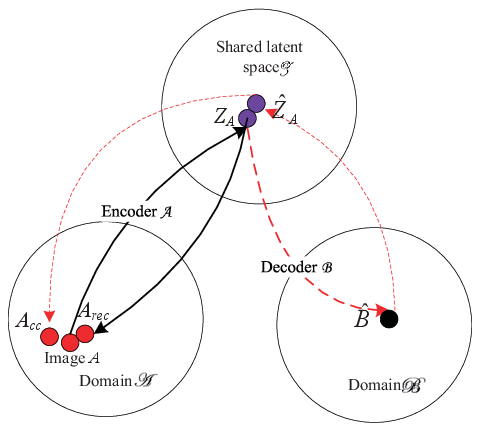}
  \caption{Cross-cycle translation between source domain, shared latent space, and target domain, taking domain $\mathscr{A}$ as the source and domain $\mathscr{B}$ as the target. The solid lines denote the processes of directly recovering $A$, and the dashed lines denote the main processes of cross-cycle translations.}
  \label{fig3}
\end{figure}

\begin{figure}
  \centering
  \includegraphics[width=0.7\linewidth]{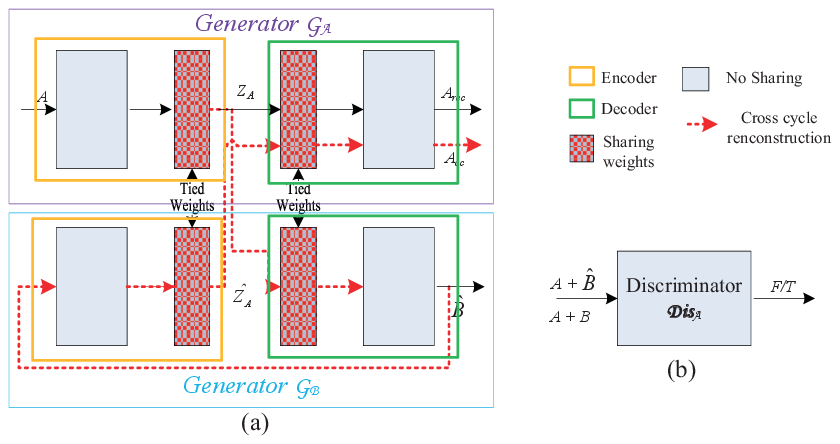}
  \caption{(a) Generator of SRUIT. Each generator consists of one encoder and one decoder. The high-level layers of both encoder and decoder are weights-shared. In brief, only the processes of translation of $A$ are depicted. (b) Discriminator of SRUIT, taken $A$ as an example. }
  \label{fig4}
\end{figure}

Based on this assumption, we tie the weights of the high layers of both generators, as Fig. \ref{fig4} shows. If there are no changes between bi-temporal images, this constraint will compel them to map into the same location, whether in latent space or target domain. When there are some changes between them, they should fall into a neighborhood of each other. Compared with the distance between images $A$ and $B$, the distance between $\hat{A}$ and $A$ is shorter and more convenient for finding their actual differences in land cover. However, the above analysis only holds in theory. There needs to be more than the weight-sharing constraint to ensure that images from two domains have the same latent code, especially unsupervised. Fortunately, bi-temporal RS images generally have many areas without changes, which will be good supervision. In addition, shared latent space allows us to implement cross-cycle translation to deal with this problem. In reality, land cover changes often happen in a unidirectional way. We can make it bidirectional in image translation by swapping the target and source domains. So, in our unsupervised translation setting, we can translate one of the bi-temporal images into the target domain and recover it. Although the cycle consistency constraint seems fit for this task, it cannot fully meet our demands of strict semantic preservation. Considering the shared latent space, we use cross-cycle consistency constraint to encourage semantics preservation in translation. Fig. \ref{fig3} illustrates cross-cycle translation from domain $\mathscr{A}$ to latent space and then to domain $\mathscr{B}$. The loss between $A_{cc}$ and $A$ will further compel $Z_A$ and $\hat{Z}_A$ to fall into the same point. In conclusion, the weight-sharing and cross-cycle consistency constraints will be mutually promoted during training to map the unchanged areas into the same location and change content into different locations in latent and target domains.

\subsection{Model}\label{sec3.2}

Fig. \ref{fig4} shows an overview of our proposed model of SRUIT and its training processes. Similar to previous works\cite{Hoffman:ICML2018,Zhu:NIPS2017,Zhu:ICCV2017,Liu:NIPS2017,Huang:ECCV2018}, the SRUIT model consists of a generator and a discriminator for each domain ($2$ domains of $\mathscr{A}$ and $\mathscr{B}$). Each generator includes two parts: an encoder and a decoder. First, the bi-temporal RS images $A$ and $B$ are fed into corresponding encoders and get their representations in the latent space. Then, the translation is performed by swapping encoder-decoder pairs. Below, we will take $A$'s processes as a sample to describe the process. Firstly, we encode image patch $A$ by encoder $\mathcal{A}$ to get its latent features $Z_A$. From here, there are two different paths. They are described as follows.
\begin{itemize}
\item The first path is directly fed $Z_A$ into decoder $\mathcal{A}$ and produces the reconstructed image $A_{rec}$. This path will promote decoder $\mathcal{A}$ to learn the prior distribution of domain $\mathscr{A}$.
\item The second path will be more complex. It will firstly feed $Z_A$ to decoder $\mathcal{B}$. Then, after decoder $\mathcal{B}$ produces $\hat{B}$, it will be sent to encoder $\mathcal{B}$ to get the recovered latent features $\hat{Z}_A$. Finally, decoder $\mathcal{A}$ receives $\hat{Z}_A$ and produces cross-cycle recovered $A_{cc}$.
\end{itemize}

Among these outputs, $A_{rec}$ and $A_{cc}$ are used to help the decoder and encoder learn related prior knowledge, and only $\hat{B}$, the translated image, is what we want. Image $A$ and $\hat{B}$ will be combined and sent to discriminator $\mathcal{D}is_\mathcal{A}$, and the expected output of the discriminator is false, while the combination of $A$ and $B$ is valid. The processes of $B$ work in a similar way.

Unlike previous works\cite{Zhu:NIPS2017,Liu:NIPS2017,You:BCCGAN}, we do not append noise vectors to the latent features, which are often used in VAE-like network structures. In addition, the prior distribution of the target domain is learned by its decoder, and the prior distribution of land covers of the source sample is carried by its latent features. These measures do restrict the diversity capacity of SRUIT to produce multimodal outputs, but this is just what we want in the name of semantic preservation.

Our loss function comprises the reconstruction loss and adversarial loss. The reconstruction loss consists of two parts: a cross-cycle reconstruction loss, which ensures that the translated result preserves enough source semantics, and a direct reconstruction loss, which encourages the decoder of each domain to learn its prior distribution. The direct reconstruction loss follows the process direction of $x\to Z_x \to x_{rec}$, $x\in\{A,B\}$.
\begin{equation}
  L_{rec}^x = E_{x\sim p(x)}[{\Vert \mathcal{D}(\mathcal{E}(x))-x\Vert}_1],
\end{equation}
where $x$ denotes the image patch sampled from the source domain, $\mathcal{E}$ is the encoder, and $\mathcal{D}$ is the decoder.

The cross-cycle reconstruction loss aims at supervising the robust semantic translation. Although it has a similar expression to the above reconstruction loss, it is generated differently. Taking $x\in\{A\}$ as an example, it follows the way of $A\to Z_A \to \hat{B} \to \hat{Z}_A \to A_{cc}$, and computed as:
\begin{equation}
  L_{cc}^x = E_{x\sim p(x)}[{\Vert \mathcal{D}_A(\mathcal{E}_B(\mathcal{D}_B(\mathcal{E}_A(x))))-x\Vert}_1].
\end{equation}
Like Pix2Pix\cite{Isola:CVPR2017}, we employ cGAN to match the distribution of translated images to the target image distribution. In order to further improve the quality of translated images, we use a multi-scale discriminator. Taking discriminator $A$ as an example, its adversarial loss is computed as
\begin{equation}
\begin{aligned}
  L_{ad}^{disa} = E_{A\sim p(A),B\sim p(B)}[log(\mathcal{D}is_A(A,B))]\\
                  + E_{A\sim p(A)}[log(1-\mathcal{D}is_A(A,\hat{B}))],
\end{aligned}
\end{equation}
where $Dis_A$ denotes the discriminator $A$, and $\hat{B}=\mathcal{D}_B(\mathcal{E}_A(A))$. The discriminator $B$ and its loss are defined similarly.

Our final objective is
\begin{equation}
\begin{split}
  G^* &=\arg\min_{\mathcal{E}_A,\mathcal{D}_A,\mathcal{E}_B,\mathcal{D}_B}\max_{\mathcal{D}is_A,\mathcal{D}is_A}\\
      &\qquad\qquad L(\mathcal{E}_A,\mathcal{D}_A,\mathcal{E}_B,\mathcal{D}_B,\mathcal{D}is_A,\mathcal{D}is_B)\\
      &=L_{ad}^{disa}+L_{ad}^{disa}+\lambda_{cc}(L_{cc}^A+L_{cc}^B)\\
      &\qquad\qquad\qquad\quad +\lambda_{rec}(L_{rec}^A+L_{rec}^B),
\end{split}
\end{equation}
where $\lambda_{cc}$ and $\lambda_{rec}$ are weights for controlling the importance of reconstruction items.

In our final objective, we do not use the loss of the latent space features as \cite{Zhu:NIPS2017,Liu:NIPS2017,You:BCCGAN} in either KL divergence or $L1$ form. The reason is that both the weights-sharing and cross-cycle consistency constraints are enough to compel the domain mapping alignment for bi-temporal images. According to our experiments, directly aligning $Z_A$ and $\hat{Z}_A$ instead distorts generators' learning and leads to worse performance.

\section{Experiments}\label{sec4}
This section demonstrates how our method (SRUIT) effectively translates paired images with semantic preservation and outperforms existing approaches. Please note that there are no good metrics for measuring semantic preservation's performance in translation. So, we will verify the validation through change detection tasks, which reflect the degree of semantic preservation. Furthermore, some other metrics reflecting the visual quality of images will also be supplied as evaluation indicators. In addition, perceptual translation of examples in Sec. \ref{sec4.2} will be given as essential supplements.

Because works related to this topic of unsupervised translation with access to paired images are rare, we compare Cycle-GAN\cite{Zhu:ICCV2017} and GC-GAN\cite{Fu:CVPR2019} for their capacities of semantic preservation. All results are generated by their official source codes after 200 epochs. GC-GAN uses rotation and flipping to ensure its best performance for a fair comparison.

\begin{figure}[h]
  \vspace{-0.1cm}
  \centering
  \subfloat{\includegraphics[width=0.4\linewidth]{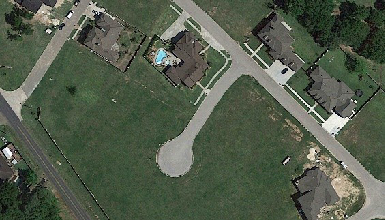}
  \label{subfig_f51}}
  \subfloat{\includegraphics[width=0.4\linewidth]{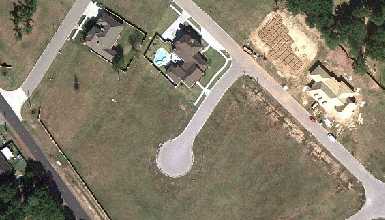}
  \label{subfig_f52}}
  \\
  \subfloat{\includegraphics[width=0.4\linewidth]{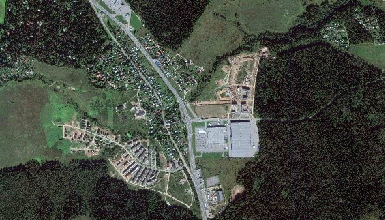}
  \label{subfig_f53}}
  \subfloat{\includegraphics[width=0.4\linewidth]{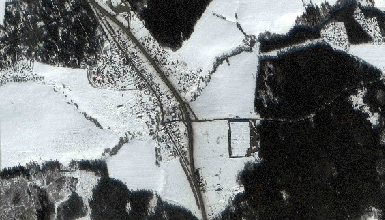}
  \label{subfig_f54}}
  \caption{The dataset CDD\cite{Lebedev:ISPRS2018} with season-varying. Images in the top row are bi-temporal images of summer and fall and have a minor difference; images in the bottom row are bi-temporal images of summer and winter and have a major difference.}
  \label{fig5}
\end{figure}

Considering that the datasets used in Pix2Pix\cite{Isola:CVPR2017} and Cycle-GAN do not fit the question here, we use bi-temporal RS images with season-varying in the CDD dataset\cite{Lebedev:ISPRS2018} to compose different datasets for evaluation. Each of the bi-temporal RS images has the size of $4725\times2700$ and will be split into patches with the size of $256\times256$ with 50\% overlap, so each domain has 756 image patches as shown in Fig. \ref{fig5}.

For the limits of paper length, more extra experimental results, including comparison with other methods, perceptual quality results, training detail, and ablation studies, are listed in the supplementary material.

\subsection{Quantitative Evaluation}\label{sec4.1}
Choosing the right metrics is critical to evaluating the translation performance quantitatively. This paper focuses on semantic preservation. Unfortunately, famous metrics such as FID\cite{Benaim:NIPS2017}, FCN-score\cite{Long:CVPR2015}, and SSIM\cite{Wang:SSIM} are not suitable and may even be misleading here, for chrominance components greatly influence them all. So, from the ultimate goal of this question studied here, we use the metrics of the change detection task to denote the semantic preservation and the distance between the translated image and the target image. The results are given in Table \ref{tab:1} and Table \ref{tab:2}. In addition, some metrics of image quality without reference (DBCNN\cite{Zhang2020DBCNN}, NIMA\cite{Hossein2018NIMA}, and total variation (TV)) are supplied in Table \ref{tab:3}. DBCNN and NIMA are blind image quality assessments using neural networks, and the higher their value, the higher the quality of the generated image. The total variation reflects the difference in details between the generated image and the original image. The smaller the value, the higher the quality of the generated image. In all tables, 'A' and 'B' denote the original images of bi-temporal image patches, 'A vs. B' means using image patches $A$ and $B$ to compose the image pair for change detection, 't-B' indicates the translated image patches from B, 'Base' means using original bi-temporal images as input data directly. GC-GAN, Cycle-GAN, and SRUIT are three methods to produce translated images. Because GC-GAN is a one-sided method, so only translated $B$ is listed.

\begin{table}
  \caption{The change detection results of translated images of bi-temporal images with minor appearance difference by three methods}
  \label{tab:1}
  \begin{tabular}{rcccccc}
    \toprule
    Method                   &Data  &Acc&mIoU&Precision&Recall&F1\\
    \midrule
    Base                     & A vs. B   & 0.988 & 0.881 & \textbf{0.956} & 0.803 & 0.873\\
    GC-GAN                   & t-B vs. B & 0.979 & 0.797 & 0.884 & 0.671 & 0.763\\
    CycleGAN                 & t-B vs. B & 0.993 & 0.930 & 0.946 & 0.913 & 0.913\\
    SRUIT                   & t-B vs. B & \textbf{0.993} & \textbf{0.937} & 0.934 & \textbf{0.939} & \textbf{0.936}\\
  \botrule
\end{tabular}
\end{table}
\begin{table}
  \caption{The change detection results of translated images of bi-temporal images with major appearance difference by three methods}
  \label{tab:2}
  \begin{tabular}{rcccccc}
    \toprule
    Method                   &Data  &Acc&mIoU&Precision&Recall&F1\\
    \midrule
    Base                     & A vs. B   & 0.959 & 0.765 & 0.769 & 0.693 & 0.729\\
    GC-GAN                   & t-B vs. B & 0.956 & 0.738 & 0.813 & 0.592 & 0.686\\
    CycleGAN                 & t-B vs. B & \textbf{0.981} & 0.879 & \textbf{0.933} & 0.824 & 0.875\\
    SRUIT                    & t-B vs. B & \textbf{0.981} & \textbf{0.880} & 0.926 & \textbf{0.833} & \textbf{0.877}\\
  \botrule
\end{tabular}
\end{table}
\begin{table}
  \caption{The loss of three methods for translated images of bi-temporal images}
  \label{tab:3}
  \begin{tabular}{r|c|ccc}
    \toprule
    Method                   &Data  &DBCNN&NIMA&TV\\
    \midrule
    SRUIT                    & t-B vs. B  & \textbf{29.20} & \textbf{3.90} & 11.95\\
    CycleGAN                 & t-B vs. B  & 25.59 & 3.85 & 13.29\\
    GC-GAN                   & t-B vs. B  & 22.87 & 3.57 & \textbf{10.13}\\
    \botrule
\end{tabular}
\end{table}

For evaluations of image perceptual quality, we use PIQ\cite{piq} to do the assessments. For change detection tasks, we construct a supervised learning model using EfficientNet\cite{Tan2019EfficientNetRM} and U-Net\cite{DBLP:Ronneberger2015}. Its source code will also be published on GitHub together with this project. Specifically, we use two EfficientNet-b1 with shared weights as the backbone and a U-Net with skip-connections as the decoder. We train the model for $300$ epochs using Adam\cite{Kingma:ICLR2015} with a learning rate of $10^{-3}$, while the learning rate decays following a 'poly' LR schedule with $0.9$. All possible flips and rotations multiple $90^\circ$ are used for data augmentations.

Our method achieves better performance than GC-GAN and Cycle-GAN in change detection metrics (Table \ref{tab:1} and Table \ref{tab:2}), while better than Cycle-GAN and worse than GC-GAN in image perceptual quality metrics (Table \ref{tab:3}). SRUIT achieves significant improvements in almost all five indexes compared to the base. In the $F1$ score, SRUIT provides $3\%$ improvement at least when applied to bi-temporal images with minor appearance differences and up to $15\%$ increase for major difference cases. This implies that our method can preserve the semantics of translated images.

\begin{figure}
  \vspace{-0.1cm}
  \centering
  \subfloat{\includegraphics[width=0.2\linewidth]{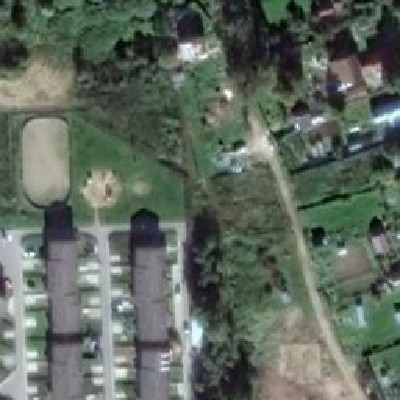}}
  \subfloat{\includegraphics[width=0.2\linewidth]{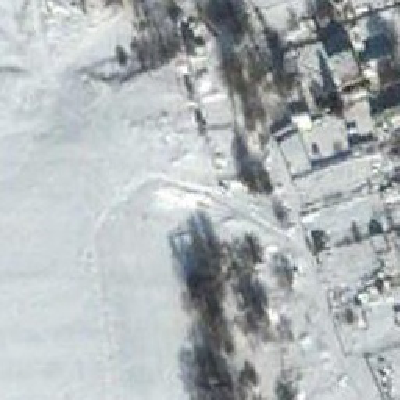}}
  \subfloat{\includegraphics[width=0.2\linewidth]{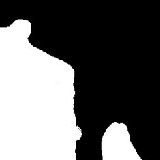}}
  \subfloat{\includegraphics[width=0.2\linewidth]{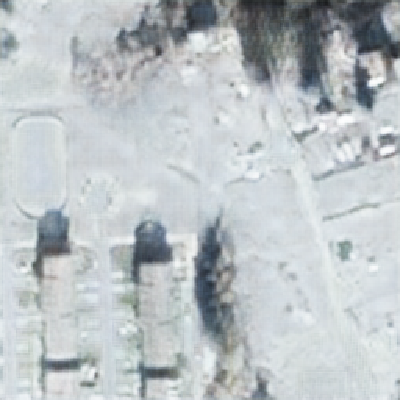}}
  \\
  \subfloat{\includegraphics[width=0.2\linewidth]{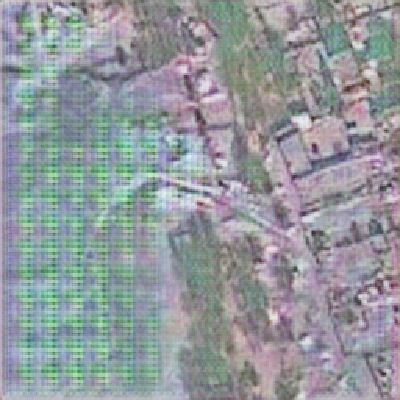}}
  \subfloat{\includegraphics[width=0.2\linewidth]{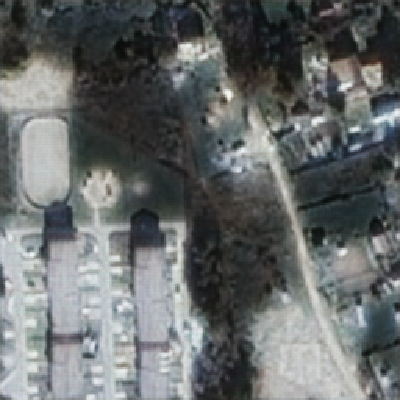}}
  \subfloat{\includegraphics[width=0.2\linewidth]{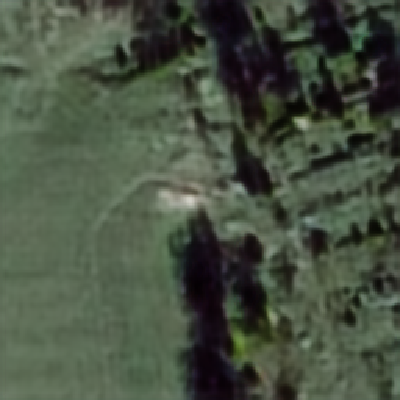}}
  \subfloat{\includegraphics[width=0.2\linewidth]{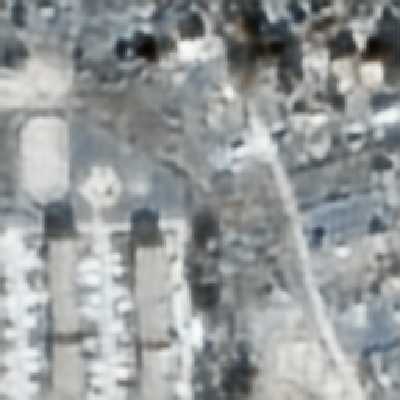}}
  \begin{center}
  \footnotesize
  (a)
  \end{center}
  \subfloat{\includegraphics[width=0.2\linewidth]{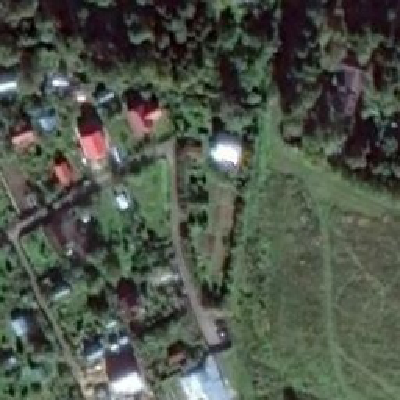}}
  \subfloat{\includegraphics[width=0.2\linewidth]{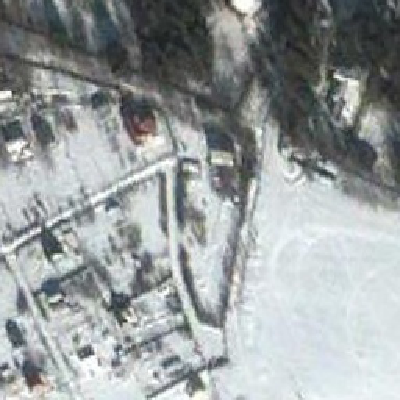}}
  \subfloat{\includegraphics[width=0.2\linewidth]{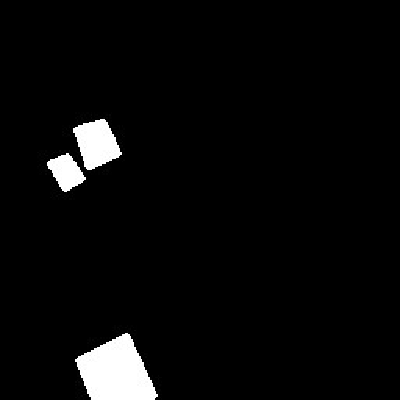}}
  \subfloat{\includegraphics[width=0.2\linewidth]{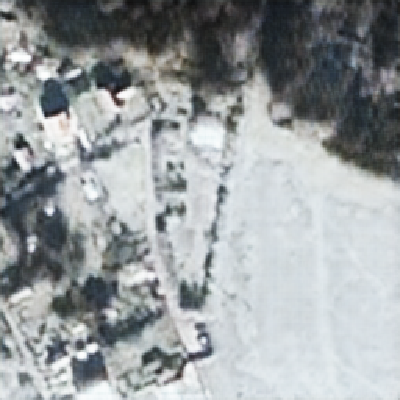}}
  \\
  \subfloat{\includegraphics[width=0.2\linewidth]{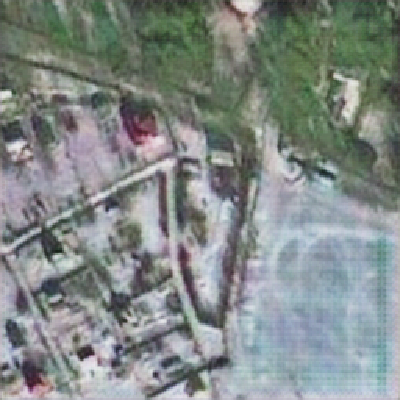}}
  \subfloat{\includegraphics[width=0.2\linewidth]{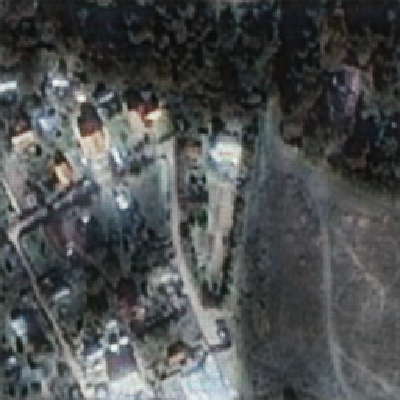}}
  \subfloat{\includegraphics[width=0.2\linewidth]{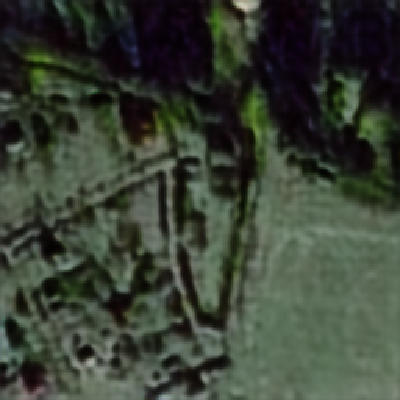}}
  \subfloat{\includegraphics[width=0.2\linewidth]{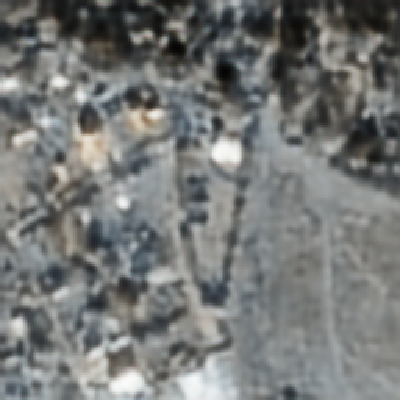}}
  \begin{center}
  \footnotesize
  (b)
  \end{center}
  \caption{The translated images of bi-temporal images with major appearance differences. (a) and (b) are selected from different images. From left to right and top to bottom are the two images $A$ and $B$ of bi-temporal images, change ground truth, translated $B$ by GC-GAN, translated $A$ and $B$ by Cycle-GAN, and translated $A$ and $B$ by ours.}
  \label{fig6}
\end{figure}

Compared with Cycle-GAN, our method has similar performance in change detection metrics. This indirectly reveals that using cycle loss does help the semantic preservation of individual images\cite{Zhu:ICCV2017,Hoffman:ICML2018}. Moreover, our method achieves better results in image perceptual quality metrics, especially for bi-temporal images with significant appearance differences. This tendency can be observed more evidently in Fig. \ref{fig6}. The results indirectly imply that using a sharing-weight constraint is helpful for stably generating unsupervised translated images with better quality when having access to paired images. Why do bad perceptual images result in good change detection results? This firstly confuses us. After several tests, we believe it is primarily because of the working mode of change detection networks. The bad perceptual images often result from noise with particular patterns. If the number of noise patterns in land covers is limited, this may lead to better classification results.

Among all methods, GC-GAN has the worst performance in change detection metrics compared with base images, while its translated image qualities have the best performance. This may be due to GC-GAN using skip connections between the encoder and decoder to improve the perceptual image quality of details. At the cost of good image quality, this may instead mislead the semantics of some regions, especially in the area of change, because too many style (chrominance) components of the target domain are integrated. This tendency can also be observed in Fig. \ref{fig6}, where some changed areas cannot be distinguished from their neighbor areas.

Overall, these quantitative evaluation results demonstrate that SRUIT can produce good perceptual images with semantic preservation. Moreover, the constraints of cross-cycle consistency and sharing weight are helpful for unsupervised stable mapping of the source image to the neighbors of the target image.

\begin{figure}
  \vspace{-0.1cm}
  \centering
  \subfloat{\includegraphics[width=0.2\linewidth]{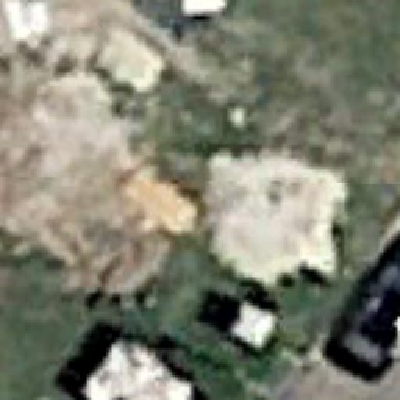}}
  \subfloat{\includegraphics[width=0.2\linewidth]{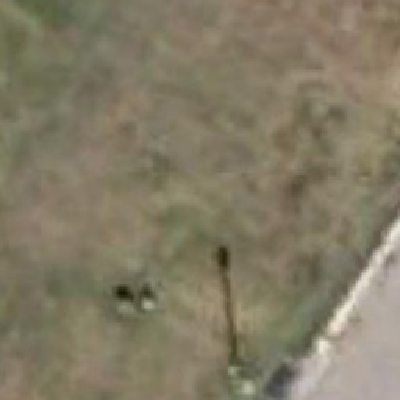}}
  \subfloat{\includegraphics[width=0.2\linewidth]{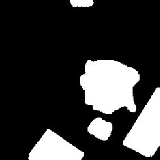}}
  \subfloat{\includegraphics[width=0.2\linewidth]{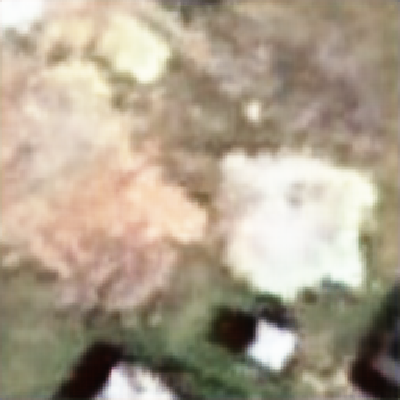}}
  \\
  \subfloat{\includegraphics[width=0.2\linewidth]{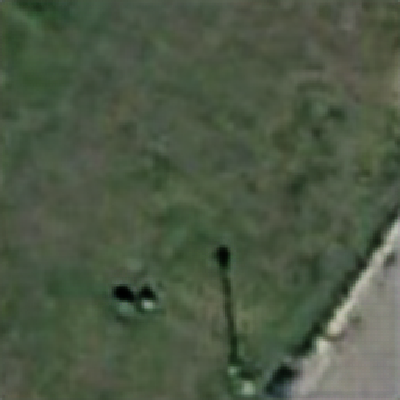}}
  \subfloat{\includegraphics[width=0.2\linewidth]{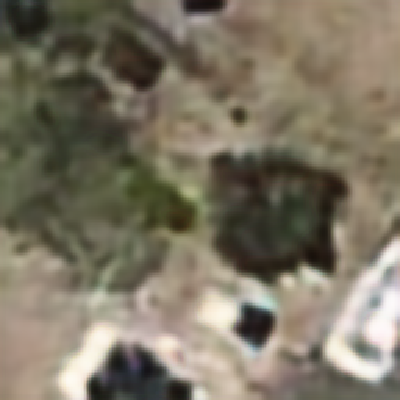}}
  \subfloat{\includegraphics[width=0.2\linewidth]{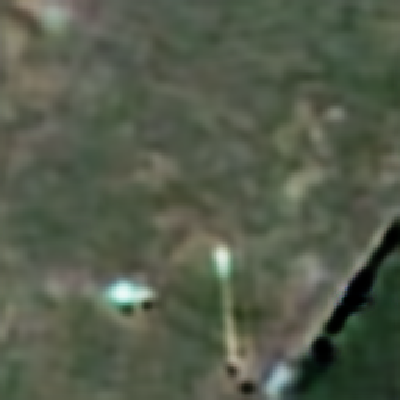}}
  \subfloat{\includegraphics[width=0.2\linewidth]{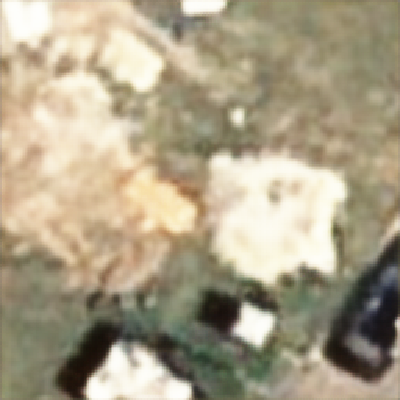}}
  \caption{The translated images of bi-temporal images with minor appearance differences. The order is the same as Figure \ref{fig6}.}
  \label{fig7}
\end{figure}

\subsection{Qualitative Evaluation}\label{sec4.2}
Besides the aforementioned quantitative tests, we show more visual results on the test datasets. Firstly, we give some examples of SRUIT in Fig. \ref{fig6} and Fig. \ref{fig7} compared to Cycle-GAN and GC-GAN. In Fig. \ref{fig7} of bi-temporal images with minor appearance differences, three methods show good performance in style transfer and semantic preservation. However, they have special performances in the case of significant appearance differences.

Translated $B$ by GC-GAN has the most similar style to image $B$. However, it is so similar that some boundaries of changed areas (e.g., the top middle areas in Fig. \ref{fig6}(b) and the top left in Fig. \ref{fig6}(a)) are washed away. The blurry boundary will harm the downstream tasks of the classification of land covers. In contrast, Cycle-GAN produces translated images with distinct area styles, e.g., the green areas in image $A$ almost translate into the dark areas in translated $B$ in Fig. \ref{fig6}(a). However, Cycle-GAN often generates translated images with strange noise patterns. In our tests, this type of image occupied almost $1/3$ of all. Two reasons may cause this. One is the difficulty of training GAN, and the other is because the source and target domains have significant deviations and too great textures. This question has vanished in Fig. \ref{fig7} of bi-temporal images with minor appearance differences. In contrast, our method does not bear this question. This contrast verifies the effectiveness of sharing weights for dealing with the studied problem.

In general, our method produces both translated $A$ and $B$ with better perceptual quality than Cycle-GAN and worse than GC-GAN, and preserves the semantics of changed areas well. Moreover, it can be noticed that the changed areas are well kept in translated images in comparison with the ground truth in both Fig. \ref{fig6} and Fig. \ref{fig7}. This demonstrates that our method is capable of semantic preservation. Especially when two translation domains have great appearance deviation, our proposed method SRUIT is superior to Cycle-GAN in image quality and GC-GAN in semantic preservation.

\begin{figure}
  \vspace{-0.1cm}
  \centering
  \subfloat[]{\includegraphics[width=0.2\linewidth]{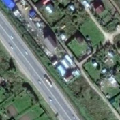}}
  \subfloat[]{\includegraphics[width=0.2\linewidth]{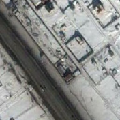}}
  \subfloat[]{\includegraphics[width=0.2\linewidth]{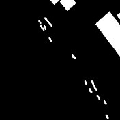}}
  \subfloat[]{\includegraphics[width=0.2\linewidth]{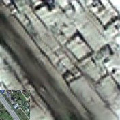}}
  \subfloat[]{\includegraphics[width=0.2\linewidth]{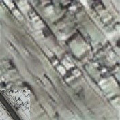}}
  \\
  \subfloat[]{\includegraphics[width=0.2\linewidth]{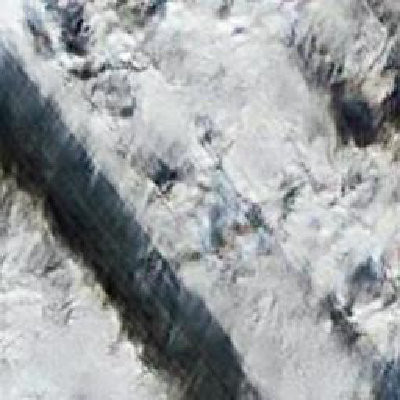}}
  \subfloat[]{\includegraphics[width=0.2\linewidth]{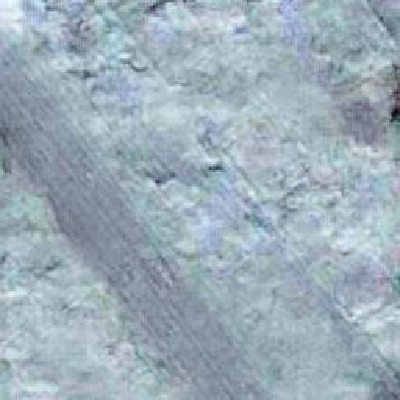}}
  \subfloat[]{\includegraphics[width=0.2\linewidth]{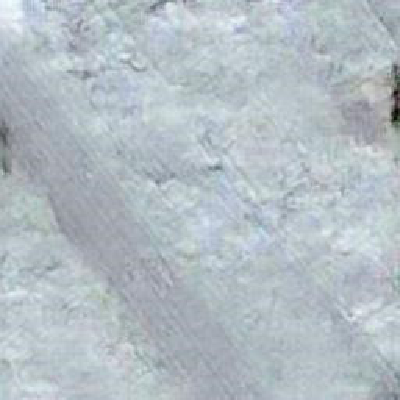}}
  \subfloat[]{\includegraphics[width=0.2\linewidth]{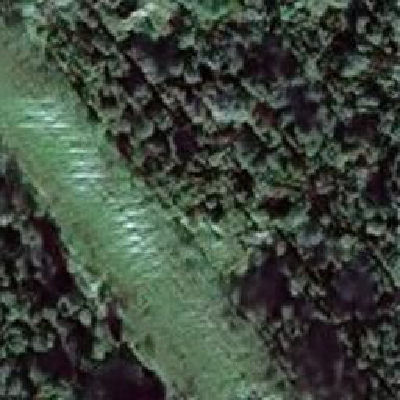}}
  \subfloat[]{\includegraphics[width=0.2\linewidth]{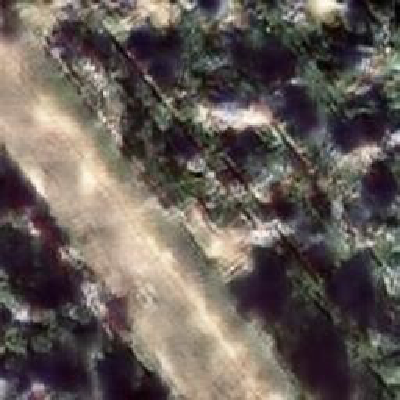}}
  \\
  \subfloat[]{\includegraphics[width=0.2\linewidth]{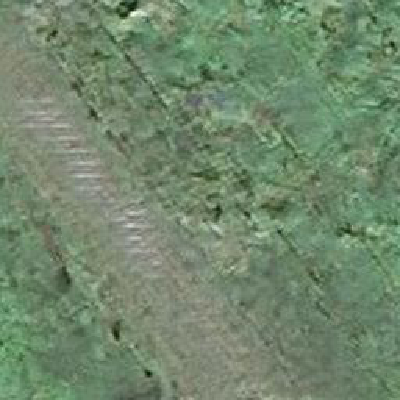}}
  \subfloat[]{\includegraphics[width=0.2\linewidth]{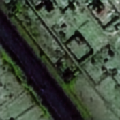}}
  \subfloat[]{\includegraphics[width=0.2\linewidth]{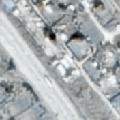}}
  \caption{Comparison with style transfer methods. (a)$\sim$(c) are bi-temporal images and ground truth, (d) is the result of AdaIN\cite{Huang:ICCV2017} taken (a) as content, (b) as style, and (e) swapping the content and style, (f)$\sim$(h) are the results of MUNIT\cite{Huang:ECCV2018} taken (a) as content, and (b) as style, and (i)$\sim$(k) swapping the content and style, and (l) and (m) are results of our methods.}
  \label{fig8}
\end{figure}

Because this work has strong relations with style transfer, we supply the comparison results with MUNIT\cite{Huang:ECCV2018} and AdaIN\cite{Huang:ICCV2017} in Fig. \ref{fig8}. In MUNIT, the model is trained on our built datasets through 100000 iterations. During the test, the style number is set to $10$, and we randomly choose three results from image $A$ to image $B$ and vice versa. MUNIT achieves far better results in summer-to-winter datasets\cite{Isola:CVPR2017,Zhu:ICCV2017,Huang:ECCV2018} than here under the same train settings. The reason may be that each domain of bi-temporal has a broader distribution. The results show that style transfer methods may lead to strange textures and are unsuitable for our study question, primarily when small changed areas exist.
\begin{figure}
  \vspace{-0.1cm}
  \centering
  \subfloat{\includegraphics[width=0.2\linewidth]{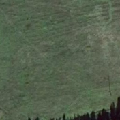}}
  \subfloat{\includegraphics[width=0.2\linewidth]{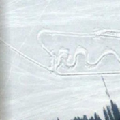}}
  \subfloat{\includegraphics[width=0.2\linewidth]{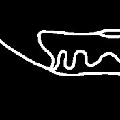}}
  \subfloat{\includegraphics[width=0.2\linewidth]{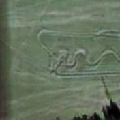}}
  \\
  \subfloat{\includegraphics[width=0.2\linewidth]{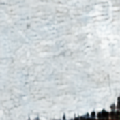}}
  \subfloat{\includegraphics[width=0.2\linewidth]{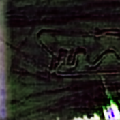}}
  \subfloat{\includegraphics[width=0.2\linewidth]{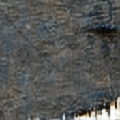}}
  \caption{Ablation of weights-sharing. From left to right and top to bottom are bi-temporal images and ground truth, two translated images of SRUIT, and two results of SRUIT dropping out of weights-sharing constraint.}
  \label{fig9}
\end{figure}
\begin{figure}
  \vspace{-0.1cm}
  \centering
  \subfloat{\includegraphics[width=0.2\linewidth]{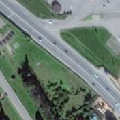}}
  \subfloat{\includegraphics[width=0.2\linewidth]{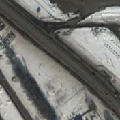}}
  \subfloat{\includegraphics[width=0.2\linewidth]{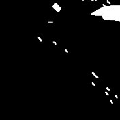}}
  \subfloat{\includegraphics[width=0.2\linewidth]{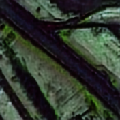}}
  \\
  \subfloat{\includegraphics[width=0.2\linewidth]{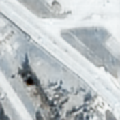}}
  \subfloat{\includegraphics[width=0.2\linewidth]{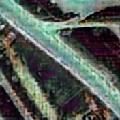}}
  \subfloat{\includegraphics[width=0.2\linewidth]{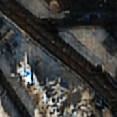}}
  \caption{Ablation of normalization layers. From left to right and top to bottom are the bi-temporal images and ground truth, two translated images of SRUIT without normalization, and two results of SRUIT with instance normalization in the discriminator.}
  \label{fig10}
\end{figure}
\begin{figure}
  \vspace{-0.1cm}
  \centering
  \subfloat{\includegraphics[width=0.2\linewidth]{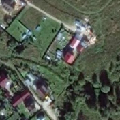}}
  \subfloat{\includegraphics[width=0.2\linewidth]{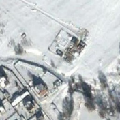}}
  \subfloat{\includegraphics[width=0.2\linewidth]{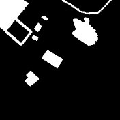}}
  \subfloat{\includegraphics[width=0.2\linewidth]{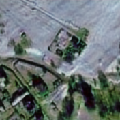}}
  \subfloat{\includegraphics[width=0.2\linewidth]{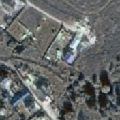}}
  \\
  \subfloat{\includegraphics[width=0.2\linewidth]{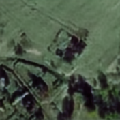}}
  \subfloat{\includegraphics[width=0.2\linewidth]{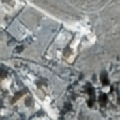}}
  \subfloat{\includegraphics[width=0.2\linewidth]{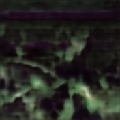}}
  \subfloat{\includegraphics[width=0.2\linewidth]{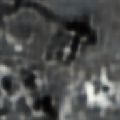}}
  \caption{Ablation of downsampling layers. From left to right and top to bottom are the bi-temporal images and ground truth, and two successive translated images of SRUIT with 3, 4, and 5 down sample layers.}
  \label{fig11}
\end{figure}

\subsection{Ablation Study}\label{sec4.3}
With the limits of paper length, we show parts of the ablation study results here. In Fig. \ref{fig9}, the results without weights-sharing only with cross-cycle-consistence constraint are given out. The results show that only cross-cycle cannot ensure high-quality translated images. The sharing-weights constraint is vital to ensure the translated image falls into the neighbors of the target domain. In other words, only using cross-cycle consistency may lead to translated images deviating from the target image style too far.

Fig. \ref{fig10} shows the ablation results of the discriminator of SRUIT w/o instance normalization. Pix2Pix and Cycle-GAN have reported that using instance instead of batch normalization layers will result in better results. However, for the question here, experimental results imply that the discriminator without normalization layers has better performance in image quality.

The proposed method uses four down-sampling operations in the generator to extract latent features. In Fig. \ref{fig11}, the results reveal that smaller down-sampling layers will not result in enough style transfer, and more will result in too blurry outcomes.

\subsection{Implementation details}\label{sec4.4}
We implement our method in the framework of Pix2Pix and Cycle-GAN. Minibatch SGD and the Adam solver\cite{Kingma:ICLR2015} are applied with a learning rate of $0.0002$, and momentum parameters $\beta_1 = 0.5$,  $\beta_2 = 0.999$. We set $\lambda_{cc}$ and $\lambda_{rec}$ in eq.(4) as $100$. The discriminator uses patch-GAN\cite{Isola:CVPR2017}, but the difference is that we use a multi-scales discriminator, and the scale is set to be $3$. The generator consists of one encoder and one decoder. The main body of the encoder consists of four $2d$ convolutions with $stride=2$, and the shared layers follow it. The shared layers include two residual convolution layers without down-sampling. Especially in the decoder of the generator, all transpose operations are implemented by up-sampling with nearest neighbors and $2d$ convolution with $stride=1$. Otherwise, the translated images will be cursed with checkboard noise. More details of the network architectures are described in the supplements.

\section{Conclusion}\label{sec5}
In this paper, we tackle the problem of unsupervised image translation with access to paired images. In essence, this problem preserves semantic information of changed areas between paired images. Taking bi-temporal RS images as examples, we build the test datasets and propose a semantically robust unsupervised image translation method. The proposed method SRUIT makes full use of the characteristics of bi-temporal RS images and achieves semantic preservation by two constraints: the weight-sharing constraint mainly for mapping two domains into the same shared latent space, and cross-cycle-consistence mainly for compelling the encoder and decoder to learn the prior distribution of each domain. The mutual promotion between the two constraints finally leads to better behaviors of semantic preservation in translation and good perceptual image quality. Quantitative and qualitative evaluations on test datasets prove that our approach can effectively preserve semantics, especially for changed areas between bi-temporal RS images.

\section*{Acknowledgements}
This work was supported in part by the Natural Science Foundation of Shandong Province under Grant ZR2022MF325 and Grant ZR2024MF113.

\section*{Declarations}

\subsection*{Competing interests}
All authors declare that they have no conflict of interest.

\subsection*{Ethics approval}
The data covered in this manuscript comes from publicly available datasets, and there are no ethical implications to discuss. All datasets involved in the current study are listed in the section of Experiments of the manuscript. This work described in this manuscript is original and has not been under consideration for publication elsewhere. This manuscript is published with the informed consent of all authors.

\subsection*{Code availability}
Code and pre-trained models are available on GitHub.com after the publication of the manuscript.

\subsection*{Author contribution}
All authors contributed to the study conception and design. Conceptualization: Sheng Fang, Jianli Zhao; Methodology: Sheng Fang, Zhe Li; Writing - original draft preparation: Sheng Fang, Kaiyu Li, Zhe Li; Writing - review and editing: Zhe Li,Xingli Zhang. All authors read and approved the final manuscript.


\bibliography{sn-bibliography}

\vspace{66pt}

\vfill

\end{document}